\pdfoutput=1
\documentclass[10pt]{cai}

\usepackage[most]{tcolorbox}
\usepackage{lmodern}
\usepackage{parskip}
\usepackage{lmodern}
\usepackage{dsfont}

\definecolor{promptaccent}{HTML}{1F77B4}

\newtcolorbox{examplebox}{
  enhanced, breakable,
  colback=gray!3,
  colframe=black!15,
  boxrule=0.4pt,
  arc=2.5pt,
  borderline west={2pt}{0pt}{black!28},
  left=10pt, right=8pt, top=4pt, bottom=4pt,
  fontupper={\footnotesize\ttfamily},
  before skip=5pt, after skip=5pt,
}

\newtcolorbox{promptbox}[1]{
  enhanced, breakable,
  colback=gray!3,
  colframe=black!15,
  boxrule=0.4pt,
  arc=2.5pt,
  borderline west={2.4pt}{0pt}{promptaccent!85!black},
  left=12pt, right=8pt, top=7pt, bottom=4pt,
  fontupper={\footnotesize\ttfamily},
  fonttitle=\bfseries\sffamily\footnotesize,
  coltitle=promptaccent!35!black,
  title=#1,
  attach boxed title to top left={xshift=8pt, yshift=-\tcboxedtitleheight/2},
  boxed title style={
    colback=white, colframe=white, boxrule=0pt,
    sharp corners, top=0pt, bottom=0pt, left=5pt, right=5pt,
  },
  before skip=6pt, after skip=5pt,
}

\definecolor{darkblue}{rgb}{0, 0, 0.5}
\hypersetup{colorlinks=true, citecolor=darkblue, linkcolor=darkblue, urlcolor=darkblue}

\definecolor{BrickRed}{HTML}{CB4154}
\definecolor{OliveGreen}{HTML}{556B2F}

\begin{document}
\begin{center}

\title{Stabilizing Black-Box Prompt Optimization with Textual Regularization and Signal Aggregation}
\maketitle

\thispagestyle{empty}

\begin{tabular}{c}
MohammadReza Davari\upstairs{\affilone \affiltwo\affilthree *},
Utkarsh Garg\upstairs{\affilthree},
Weixin Cai\upstairs{\affilthree},
Eugene Belilovsky\upstairs{\affilone \affiltwo}
\\[0.25ex]
{\small \upstairs{\affilone} Concordia University} \quad
{\small \upstairs{\affiltwo} Mila -- Quebec AI Institute} \quad
{\small \upstairs{\affilthree} Microsoft} \\
\end{tabular}

\emails{
  \upstairs{*}mohammadreza.davari@concordia.ca
}
\vspace*{0.2in}
\end{center}

\begin{abstract}
An increasing number of NLP applications interact with large language models (LLMs) through black-box APIs, making prompt engineering critical for controlling model behavior. Recent Automatic Prompt Optimization (APO) methods iteratively refine prompts using model-generated critiques (often called as \emph{textual gradients}), but they predominantly optimize from failures and underutilize information contained in correct predictions, leading to instability and semantic drift. We propose \textbf{TRAS} (Textual Regularization with Aggregated Signals), a feedback-centric framework that is plug-and-play with existing APO search backbones. It retains the standard textual gradient signal from prior work for error correction, and introduces a complementary \emph{textual regularizer} derived from successful predictions to preserve beneficial prompt components. Because both signals are stochastic and can be noisy, we further introduce \emph{Monte Carlo Signal Aggregation (MCSA)}, which samples multiple gradients or regularizers and aggregates them into a single actionable directive, emphasizing consistent, actionable advice while filtering out outliers. Motivated by rapid model churn, we also formalize \emph{Automatic Prompt Migration (APM)}, the practical problem of adapting an expert prompt across model versions or API providers without losing critical instructions. Across standard APO and APM scenarios, our approach consistently outperforms strong baselines, yielding higher accuracy, faster convergence, and lower query cost, while substantially reducing the degradation observed under naive prompt migration.\footnote{Code is available at \url{https://github.com/rezazzr/TRAS}.}
\end{abstract}


\section{Introduction}

\begin{figure}[t]
  \centering
  \includegraphics[width=0.90\columnwidth]{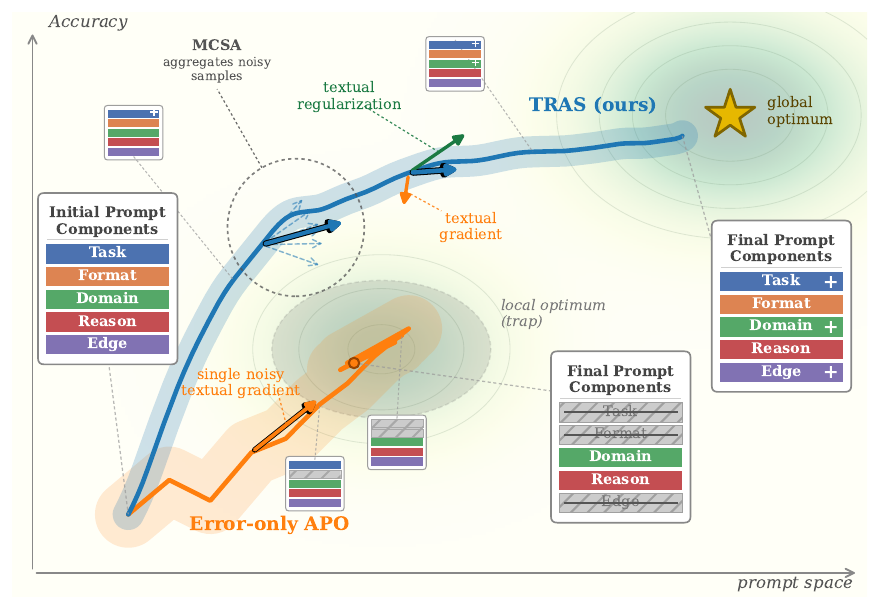}
  \caption{TRAS preserves prompt structure across optimization.
  Error-only APO (orange) updates prompts from failures alone; the
  trajectory oscillates with a wide variance band and working prompt
  components get erased (\emph{instruction loss}). TRAS (blue) adds a
  textual regularizer from successes and aggregates multiple samples
  per signal (MCSA), producing a smoother trajectory with a narrower
  variance band that better approaches the global optimum while preserving and
  refining components. Illustrative of the APM regime
  (\texttt{GPT-3.5-turbo}$\to$\texttt{GPT-4o}); full results in
  Table~\ref{tab:prompt_migration_results}.}
  \label{fig:teaser}
\end{figure}

Traditionally, NLP tasks have relied on fine-tuning pretrained models~\cite{bommasani2021opportunities,devlin2019bert,lewis-etal-2020-bart,radford2018improving,raffel2020exploring} on downstream datasets~\cite{davari2019toponym,farahnak2021semantic,davari2020timbert,yang2023toxbuster,marks2024clac,davari2020neural}, leveraging internal representations such as hidden states~\cite{rogers2021primer,davari2022probing}, gradients, and attention patterns~\cite{kornblith2019similarity,raghu2021vision,davari2023reliability,davari2022inadequacy} to drive methods like prompt tuning~\cite{li2021prefix,lester2021power}, LoRA~\cite{hu2022lora}, and other parameter-efficient approaches~\cite{davari2024model-a,yadav2023ties,davari2024model,yu2024language}.
However, the landscape is increasingly shifting toward closed-weight LLMs accessed via black-box APIs~\cite{ai2023gpt,bubeck2023sparks}, where internal representations are unavailable and prompt design becomes the primary mechanism for controlling model behavior~\cite{luo2022biogpt,kojima2022large,wang2022self,zhou2022least,madaan2023self,bai2022constitutional,chen2023teaching}. Manual prompt engineering remains costly, requiring domain expertise and substantial trial-and-error~\cite{jiang2023structgpt,wei2022chain,kong2023better}.

Recent \emph{Automatic Prompt Optimization} (APO) methods~\cite{wang2023promptagent,yang2023large,zhou2022large,pryzant2023automatic} iteratively refine prompts using feedback from model behavior. In many such approaches, the dominant signal is an LLM-generated critique from incorrect predictions, a \emph{textual gradient}~\cite{wang2023promptagent,pryzant2023automatic}. While effective for error correction, this view underutilizes information from successful predictions and can lead to \emph{semantic drift}: edits that fix local failures may inadvertently degrade globally useful prompt components (Figure~\ref{fig:teaser}). Importantly, correct predictions carry information about \emph{which parts of a prompt are working}, and ignoring this signal makes optimization less stable.

In this paper, we focus on the quality and reliability of the textual update signal, an axis orthogonal to the choice of search strategy~\cite{wang2023promptagent,pryzant2023automatic,schnabel2024sammo} (which selects \emph{which} prompt to refine). We introduce a complementary \emph{textual regularization} signal derived from successes that encourages preserving beneficial prompt components, and \emph{Monte Carlo Signal Aggregation (MCSA)} which samples multiple independent signals and aggregates them to reduce variance while managing dilution.

Prompt optimization becomes especially challenging under rapid model churn, where practitioners must migrate optimized prompts across model versions or providers. We formalize this as \emph{Automatic Prompt Migration (APM)}~\cite{library995684}: adapting an expert source prompt to a target model while avoiding \emph{instruction loss}, the inadvertent deletion of load-bearing prompt components during re-optimization~\cite{kojima2022large,zhou2022large,zhang2022tempera,ma2023let,chen2023you}. This mirrors the \emph{continual learning} setting of adapting to new tasks while preserving prior knowledge without access to past data~\cite{asadi2023prototype}, and motivates a regularization-centric update rule rather than aggressive re-optimization. Our method targets this failure mode by using textual regularization to preserve transferable structure and MCSA to reduce noisy updates.

We combine these ideas into \textbf{TRAS} (\textbf{T}extual \textbf{R}egularization with \textbf{A}ggregated \textbf{S}ignals), a unified framework for robust black-box prompt optimization and migration that consistently improves accuracy, accelerates convergence, and reduces API usage across standard APO and APM scenarios.

Our contributions are:
\begin{enumerate}
  \item We formalize \emph{Automatic Prompt Migration (APM)} as a practical black-box setting, highlighting instruction loss as a key failure mode of na\"ive migration and re-optimization.
  \item We propose \textbf{TRAS}, augmenting textual-gradient-based APO with (i) \emph{textual regularization} from successes to stabilize updates and (ii) \emph{Monte Carlo Signal Aggregation} to reduce signal variance while managing dilution, yielding robust gains in both APO and APM.
\end{enumerate}

\begin{wrapfigure}[32]{r}{0.40\textwidth}
  \vspace{-0em}
  \centering
  \vspace{0em}
  \includegraphics[width=\linewidth]{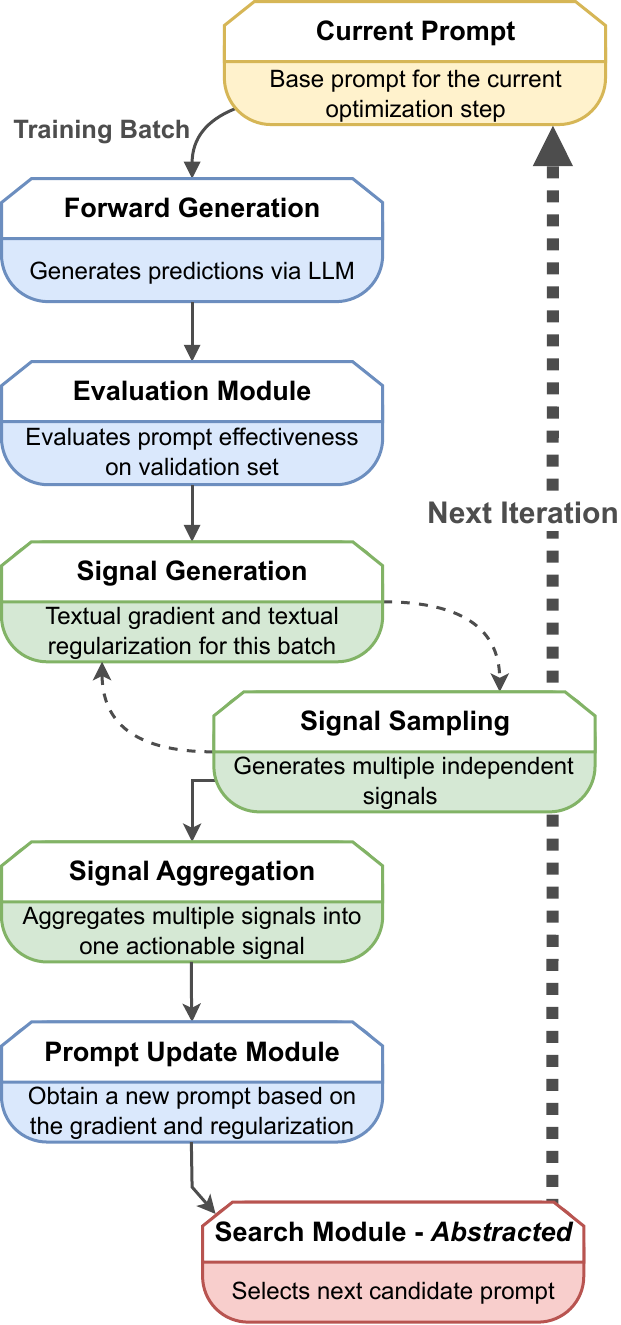}
  \vspace{0em}
  \caption{Overview of our proposed framework for automatic prompt optimization (APO). The framework consists of five primary modules. The Search Module is abstracted to allow for the integration of various search and planning methods.}
  \label{fig:framework}
\end{wrapfigure}

\section{Related Work}

APO methods can be broadly categorized by their level of access to model internals. Methods with full or partial access to parameters, gradients, or output probabilities, applicable to open-source models like LLaMA~\cite{touvron2023llama,touvron2023llama2,grattafiori2024llama} or Mistral~\cite{jiang2023mistral,mistral2024nemo}, can train soft prompts~\cite{li2021prefix,lester2021power,hu2022lora,wang2023multitask,qin2021learning} or optimize discrete prompts via gradient-guided search~\cite{shin2020autoprompt,wen2023hard,gao2020making,chen2023instructzero,hao2023optimizing,wang2022self}. These techniques are infeasible for black-box APIs, our primary focus.

Black-box APO methods fall into two families. \emph{Iterative generation} methods (e.g., APE~\cite{zhou2022large}, OPRO~\cite{yang2023large}) repeatedly propose prompt candidates, evaluate them on held-out data, and select top performers. These methods rely on \emph{scalar feedback} (e.g., accuracy) and do not explicitly represent \emph{why} a prompt fails or succeeds. \emph{Search and planning} methods (e.g., PromptAgent~\cite{wang2023promptagent}, ProTeGi~\cite{pryzant2023automatic}, SAMMO~\cite{schnabel2024sammo}) frame APO as structured exploration, using MCTS~\cite{coulom2006efficient}, beam search, or multi-objective optimization, guided by \emph{textual gradients} derived from failures. These provide stronger exploration, but the inner-loop update signal is derived primarily from errors, which is inherently stochastic and can contribute to semantic drift.
Our work complements these methods by improving the reliability and informativeness of the update signal, while remaining plug-and-play with different search strategies.

\section{Method}
\label{sec:method}

We study APO for black-box LLMs accessed via APIs. A task is defined by $\mathcal{T}=(\mathcal{X},\mathcal{Y},\mathcal{D},\ell)$, where $\mathcal{X}$ is the input space, $\mathcal{Y}$ the output space, $\mathcal{D}$ a distribution over $(x,y)\in\mathcal{X}\times\mathcal{Y}$, and $\ell$ a task loss. Let $\mathcal{P}$ denote the discrete prompt space; we write $p \in \mathcal{P}$ for an individual prompt. A black-box LLM induces $\hat{y} \sim P_{\theta}(\cdot \mid x,p)$, where we assume no access to $\theta$, internal activations, or gradients. The APO objective is to find:
\begin{equation}
p^\star \in \arg\min_{p\in\mathcal{P}} R(p),
\qquad
R(p) := \mathbb{E}_{(x,y)\sim\mathcal{D}}
\Big[
\mathbb{E}_{\hat{y}\sim P_\theta(\cdot\mid x,p)}
\big[\ell(\hat{y},y)\big]
\Big].
\label{eq:risk}
\end{equation}
In practice, the risk $R(p)$ is estimated on a finite validation set. Under $0$--$1$ loss, minimizing risk is equivalent to maximizing accuracy.

Our framework is modular and can be combined with existing search backbones; it focuses on improving the \emph{textual signals} used to revise prompts. The overall pipeline is shown in Figure~\ref{fig:framework}. At each step $t$, given a prompt $p_t$, we run it on a batch, partition outcomes into successes and failures, generate corrective and stabilizing signals, aggregate them via MCSA, and rewrite the prompt. We detail each component below.

\subsection{Forward Generation and Evaluation}
\label{sec:method_eval}

At step $t$, we sample a mini-batch $B_t=\{(x_i,y_i)\}_{i=1}^n$ from the training set and obtain predictions $\hat{y}_i$ from $P_M(\cdot\mid x_i,p_t)$. We partition the batch by correctness $c_i = \mathds{1}[\hat{y}_i = y_i]$ into an error subset $B_t^{-}$ and a success subset $B_t^{+}$.

\subsection{Textual Gradient}
\label{sec:textual_gradient}
Following prior work~\cite{pryzant2023automatic, wang2023promptagent}, we derive a corrective signal from incorrect predictions. A critique policy $\pi_g$ produces a natural-language signal $g_t \sim \pi_g(\cdot \mid p_t, B_t^{-})$ that identifies \emph{what to change} (e.g., missing constraints, ambiguous instructions, formatting issues) to fix observed failures.

\subsection{Textual Regularization}
\label{sec:textual_regularization}
Error-driven updates alone can cause \emph{semantic drift}: edits that fix local failures may delete globally beneficial prompt components, forcing the optimizer into costly \emph{remove-and-rediscover} cycles where useful instructions are first overwritten and must later be re-learned.
To stabilize optimization, we introduce a success-conditioned \emph{textual regularization} signal that prescribes \emph{preservation} rather than mutation.
An attribution policy $\pi_r$ analyzes successes and returns a regularizer $r_t \sim \pi_r(\cdot \mid p_t, B_t^{+})$, which attributes correctness to specific prompt components and expresses \emph{do-not-remove / keep-unchanged / strengthen} constraints. To avoid anchoring early iterations to a poor initialization, we employ a warm-up schedule:
\begin{equation}
r_t :=
\begin{cases}
\varnothing, & t < \tau_{\text{warmup}},\\
\text{sample from }\pi_r(\cdot \mid p_t, B_t^{+}), & t \ge \tau_{\text{warmup}},
\end{cases}
\label{eq:reg_schedule}
\end{equation}
so the optimizer first explores using only corrective signals, then refines while preserving verified structure.

\subsection{Monte Carlo Signal Aggregation (MCSA)}
\label{sec:mcsa}
Both $g_t$ and $r_t$ are stochastic LLM outputs and can have high variance. We reduce variance by sampling $K$ independent signals and aggregating them:
\begin{align}
g_{t,1},\dots,g_{t,K} &\overset{\text{i.i.d.}}{\sim} \pi_g(\cdot \mid p_t, B_t^{-}), \label{eq:mc_g_samples}\\
r_{t,1},\dots,r_{t,K} &\overset{\text{i.i.d.}}{\sim} \pi_r(\cdot \mid p_t, B_t^{+}). \label{eq:mc_r_samples}
\end{align}
An aggregation operator (LLM summarizer) combines samples into consolidated directives:
\begin{equation}
\bar{g}_t := \Phi_g(\{g_{t,k}\}_{k=1}^K),
\qquad
\bar{r}_t := \Phi_r(\{r_{t,k}\}_{k=1}^K).
\label{eq:mc_aggregate}
\end{equation}
The aggregated signals emphasize consistent, actionable edits and de-emphasize outliers. Increasing $K$ reduces sampling variance but can increase dilution when too many signals are compressed into a fixed-length directive, producing generic summaries. This variance--dilution trade-off yields a sweet spot at moderate $K$, which we characterize empirically in Section~\ref{sec:ablation}.

\subsection{Prompt Update}
\label{sec:update_operator}
Given $p_t$ and aggregated signals $(\bar{g}_t,\bar{r}_t)$, we generate the next prompt via an update policy:
\begin{equation}
p_{t+1} \sim \pi_{\text{upd}}(\cdot \mid p_t, \bar{g}_t, \bar{r}_t, B_t),
\label{eq:update_rule}
\end{equation}
which applies corrective changes from $\bar{g}_t$ while preserving components identified by $\bar{r}_t$. TRAS is orthogonal to the upstream search strategy that selects which prompt $p_t$ to refine; in our experiments we use the MCTS strategy that was used in PromptAgent~\cite{wang2023promptagent} as the search backbone due to its established effectiveness~\cite{zhang2025mars, li2025survey}.

\subsection{TRAS for Automatic Prompt Migration}
\label{sec:apm}
We formalize \emph{prompt migration} as adapting an expert prompt $p_S^\star$, optimized for a source LLM $P_{\theta_S}$ (e.g., \texttt{GPT-3.5-turbo}), to a target LLM $P_{\theta_T}$ (e.g., \texttt{GPT-4o}):
\begin{equation}
\min_{p\in\mathcal{P}} R_T(p)
\quad \text{s.t.} \quad
p \in \mathcal{N}(p_S^\star),
\label{eq:trust-region-problem}
\end{equation}
where $R_T$ is the target-model risk and $\mathcal{N}(\cdot)$ encodes preservation of successful prompt structure. Na\"ive error-driven updates can quickly overwrite load-bearing instructions in $p_S^\star$. We initialize $p_0 := p_S^\star$ and activate textual regularization \emph{immediately} ($\tau_{\text{warmup}} := 0$), so that $r_t$ is sampled for all $t \ge 0$. This allows TRAS to correct target-specific mismatches via $\bar{g}_t$ while preserving the transferable structure of $p_S^\star$ via $\bar{r}_t$. We also use a smaller MCSA sample count $K$ in this regime (see Section~\ref{sec:prompt_migration}) because the warm-started expert prompt produces lower-variance signals than the cold-start APO setting.

\section{Experiments}
\label{sec:experiments}

We evaluate TRAS on five reasoning tasks spanning causal, spatial, tabular, entailment, and semantic similarity, using \texttt{GPT-3.5-turbo} (APO) and \texttt{GPT-4o} (APM on GPT-3.5-turbo to GPT-4o).
We optimize prompts on validation sets and report final performance on test sets; the primary metric is accuracy.
Full details on tasks, dataset statistics, splits, and default prompts are provided in Appendix~\ref{sec:appendix-dataset-statistics} and Appendix~\ref{sec:appendix-base-prompt}.

\subsection{Standard Prompt Optimization (GPT-3.5-turbo)}
\label{sec:prompt_tuning}
We evaluate TRAS in the standard APO setting using \texttt{GPT-3.5-turbo}. Each run is initialized with a task-specific default prompt (Appendix~\ref{sec:appendix-base-prompt}) and run for up to 15 iterations using the PromptAgent~\cite{wang2023promptagent} search backbone.
MCSA samples $K=6$ signals per batch, and textual regularization activates after a short warm-up ($\tau_{\text{warmup}}=3$ or $4$ depending on the dataset; see Section~\ref{sec:ablation}).
Table~\ref{tab:prompt_tuning_results} reports results averaged over five seeds.
TRAS yields consistent and statistically significant accuracy improvements across all tasks, ranging from $4.9\%$ to $21.5\%$ over PromptAgent ($p<0.01$).
MCSA reduces standard deviation across seeds, highlighting the role of aggregated signals in stabilizing optimization.
TRAS also reduces LLM calls by $0.5\%$ to $3.3\%$ relative to the baseline by avoiding unnecessary removal-and-rediscovery cycles (Section~\ref{sec:appendix-experimentation-costs}).

\begin{table}[tb]
  \centering
  \begin{tabular}{l l c r r}
    \toprule
    \textbf{Dataset (Init. Acc.)} & \textbf{Method}                     & \textbf{Accuracy}        & \textbf{$p$-value} & \textbf{Cohen's $d$} \\
    \midrule
    \multirow{4}{*}{%
    \shortstack[l]{Causal Judgment                                                                                                             \\(56.5 {\scriptsize{± 3.67}} )}%
    }
                                  & Baseline                            & 58.6 \scriptsize{± 3.98} & -                  & -                    \\
                                  & Baseline + MCSA\textsuperscript{*}
                                  & 60.8 \scriptsize{± 2.38}            & 0.020                    & 1.687                                     \\
                                  & Baseline + TR\textsuperscript{*}
                                  & 63.6 \scriptsize{± 2.62}            & 0.040                    & 1.336                                     \\
                                  & TRAS\textsuperscript{**}
                                  & \textbf{64.4} {\scriptsize{± 2.16}} & 0.008                    & 2.162                                     \\
    \midrule
    \multirow{4}{*}{%
    \shortstack[l]{Geometric Shapes                                                                                                            \\(32.7 {\scriptsize{± 2.04}} )}%
    }
                                  & Baseline                            & 52.1 \scriptsize{± 4.94} & -                  & -                    \\
                                  & Baseline + MCSA\textsuperscript{*}
                                  & 57.8 \scriptsize{± 3.15}            & 0.032                    & 1.439                                     \\
                                  & Baseline + TR\textsuperscript{*}
                                  & 61.6 \scriptsize{± 4.18}            & 0.035                    & 1.399                                     \\
                                  & TRAS\textsuperscript{**}
                                  & \textbf{63.3} {\scriptsize{± 1.16}} & 0.004                    & 2.644                                     \\
    \midrule
    \multirow{4}{*}{%
    \shortstack[l]{Penguins                                                                                                                    \\(60.5 {\scriptsize{± 4.87}} )}%
    }
                                  & Baseline                            & 65.1 \scriptsize{± 4.96} & -                  & -                    \\
                                  & Baseline + MCSA\textsuperscript{*}
                                  & 66.1 \scriptsize{± 2.42}            & 0.083                    & 1.148                                     \\
                                  & Baseline + TR\textsuperscript{*}
                                  & 66.9 \scriptsize{± 3.97}            & 0.043                    & 1.464                                     \\
                                  & TRAS\textsuperscript{**}
                                  & \textbf{68.6} {\scriptsize{± 1.87}} & 0.007                    & 2.556                                     \\
    \midrule
    \multirow{4}{*}{%
    \shortstack[l]{Biosses                                                                                                                     \\(25.2 {\scriptsize{± 3.84}} )}%
    }
                                  & Baseline                            & 62.5 \scriptsize{± 4.19} & -                  & -                    \\
                                  & Baseline + MCSA\textsuperscript{*}
                                  & 67.0 \scriptsize{± 2.92}            & 0.044                    & 1.456                                     \\
                                  & Baseline + TR\textsuperscript{*}
                                  & 68.2 \scriptsize{± 3.02}            & 0.021                    & 1.844                                     \\
                                  & TRAS\textsuperscript{**}
                                  & \textbf{70.4} {\scriptsize{± 2.02}} & 0.006                    & 2.654                                     \\
    \midrule
    \multirow{4}{*}{%
    \shortstack[l]{CB                                                                                                                          \\(68.5 {\scriptsize{± 4.22}} )}%
    }
                                  & Baseline                            & 81.7 \scriptsize{± 3.17} & -                  & -                    \\
                                  & Baseline + MCSA\textsuperscript{*}
                                  & 84.2 \scriptsize{± 2.02}            & 0.032                    & 1.610                                     \\
                                  & Baseline + TR\textsuperscript{*}
                                  & 84.2 \scriptsize{± 3.73}            & 0.049                    & 1.402                                     \\
                                  & TRAS\textsuperscript{**}
                                  & \textbf{85.7} {\scriptsize{± 3.54}} & 0.008                    & 2.495                                     \\
    \bottomrule
  \end{tabular}
\caption{
  Prompt optimization on \texttt{GPT-3.5-turbo} (mean accuracy $\pm$ std., five seeds).
  Significance vs.\ PromptAgent is via paired $t$-tests ($p$, Cohen's $d$).
  +MCSA: Monte Carlo Signal Aggregation ($K{=}6$); +TR: Textual Regularization; TRAS: both.
  Init.\ Acc.: default prompt.
  All methods run 15 iterations; +TR activates at $\tau_{\text{warmup}}{=}3$ (Causal Judgment/Geometric Shapes/Penguins) or $4$ (Biosses/CB).
  TRAS improves accuracy and reduces variance over baselines.
  }
  \label{tab:prompt_tuning_results}
\end{table}

\subsection{Prompt Migration: GPT-3.5-turbo \textrightarrow GPT-4o}
\label{sec:prompt_migration}

We evaluate TRAS in the APM setting, where expert prompts optimized on \texttt{GPT-3.5-turbo} initialize optimization on \texttt{GPT-4o}.
Throughout this subsection, \emph{DP} (Default Prompt) refers to the task-specific default prompt used as a cold-start initializer (see Appendix~\ref{sec:appendix-base-prompt}), and \emph{EP} (Expert Prompt) refers to a prompt already optimized on \texttt{GPT-3.5-turbo} that we transfer as the initializer on \texttt{GPT-4o}.
As shown in Table~\ref{tab:gpt4_migration_base_study}, direct transfer of expert prompts yields initial gains of $1.3\%$--$3.3\%$ over default prompts, but na\"ive re-optimization via PromptAgent often erodes these gains due to instruction loss, early corrective signals overwrite useful instructions embedded in the expert prompt.

\begin{table}[tb]
  \centering
  \begin{tabular}{l l c c l r}
    \toprule
    \textbf{Dataset}
     & \textbf{Stage}
     & \textbf{DP Acc.}
     & \textbf{EP Acc.}
     & \textbf{$p$-value}
     & \textbf{Cohen's $d$}                                                                                                   \\
    \midrule
    \multirow{2}{*}{Causal Judgment}
     & Initial              & 71.8 {\scriptsize ± 1.92}          & \textbf{74.2} {\scriptsize ± 3.46} & 0.033$^{*}$  & 1.434  \\
     & Final                & 73.4 {\scriptsize ± 1.82}          & {\bf 73.8} {\scriptsize ± 1.79}    & 0.803        & 0.119  \\
    \midrule
    \multirow{2}{*}{Geometric Shapes}
     & Initial              & 54.8 {\scriptsize ± 1.89}          & \textbf{58.2} {\scriptsize ± 2.22} & 0.001$^{**}$ & 4.138  \\
     & Final                & \textbf{79.0} {\scriptsize ± 5.32} & 75.1 {\scriptsize ± 5.80}          & 0.040$^{*}$  & -1.344 \\
    \midrule
    \multirow{2}{*}{Penguins}
     & Initial              & 92.9 {\scriptsize ± 1.85}          & \textbf{95.8} {\scriptsize ± 1.72} & 0.025$^{*}$  & 1.745  \\
     & Final                & \textbf{95.2} {\scriptsize ± 2.03} & 92.3 {\scriptsize ± 2.89}          & 0.005$^{**}$ & -2.771 \\
    \midrule
    \multirow{2}{*}{Biosses}
     & Initial              & 69.9 {\scriptsize ± 2.73}          & \textbf{72.2} {\scriptsize ± 1.78} & 0.0125$^{*}$ & 2.159  \\
     & Final                & \textbf{76.7} {\scriptsize ± 2.84} & 76.1 {\scriptsize ± 1.97}          & 0.381        & -0.600 \\
    \midrule
    \multirow{2}{*}{CB}
     & Initial              & 79.3 {\scriptsize ± 1.57}          & \textbf{80.3} {\scriptsize ± 1.54} & 0.005$^{**}$ & 2.855  \\
     & Final                & \textbf{80.0} {\scriptsize ± 4.62} & 78.7 {\scriptsize ± 2.13}          & 0.381        & -0.492 \\
    \bottomrule
  \end{tabular}
    \caption{
    Transferability of expert prompts (EP) from \texttt{GPT-3.5-turbo} to \texttt{GPT-4o} (accuracy mean $\pm$ std., five seeds).
    We compare Default Prompt (DP) vs.\ transferred EP at \textit{Initial} (direct transfer) and \textit{Final} (after 15 PromptAgent iterations on \texttt{GPT-4o}).
    Significance via paired $t$-tests.
    Direct transfer yields modest gains ($1.3$--$3.3$\%), but na\"ive re-optimization with PromptAgent often erodes them (instruction loss).
    }
  \label{tab:gpt4_migration_base_study}
\end{table}

We next evaluate TRAS for APM. TRAS activates textual regularization from the first iteration ($\tau_{\text{warmup}}=0$) and uses $K=2$ for MCSA. Two factors motivate the smaller $K$: (i) the warm-started expert prompt is already strong, so per-iteration signals are less variable than in the cold-start APO setting and pilot runs showed no accuracy gain from $K\ge 3$; and (ii) \texttt{GPT-4o} API costs scale with $K$, so larger sweeps were not tractable at this budget.
Table~\ref{tab:prompt_migration_results} summarizes results.
TRAS consistently improves performance across all tasks, with accuracy gains of $3.5\%$ to $16.0\%$ over the baseline ($p<0.01$).
The largest gains occur in Geometric Shapes ($12.5\%$) and Biosses ($16.0\%$), where preserving domain-specific structure is especially important.
TRAS also reduces total LLM calls by $4.2\%$ to $6.2\%$ (Section~\ref{sec:appendix-experimentation-costs}).

\begin{table}[tb]
    \centering
  \begin{tabular}{l l c r r}
    \toprule
    \textbf{Dataset (Init. Acc.)}
     & \textbf{Method}
     & \textbf{Final Acc.}
     & \textbf{$p$-value}
     & \textbf{Cohen's $d$}                                                                \\
    \midrule
    \multirow{4}{*}{\shortstack[l]{%
    Causal Judgment                                                                        \\
    DP: 71.8 {\scriptsize ± 1.92}                                                          \\
        EP: 74.2 {\scriptsize ± 3.46}
      }}
     & Baseline                         & 73.8 {\scriptsize ± 1.79}       & -      & -     \\
     & Baseline + MCSA\textsuperscript{*} & 74.7 {\scriptsize ± 1.44}       & 0.053  & 1.214 \\
     & Baseline + TR\textsuperscript{*} & 75.8 {\scriptsize ± 1.78}       & 0.047  & 1.265 \\
     & TRAS\textsuperscript{**}        & {\bf 76.4} {\scriptsize ± 1.59} & 0.007  & 2.280 \\
    \midrule
    \multirow{4}{*}{\shortstack[l]{%
    Geometric Shapes                                                                       \\
    DP: 54.8 {\scriptsize ± 1.89}                                                          \\
        EP: 58.2 {\scriptsize ± 2.22}
      }}
     & Baseline                         & 75.1 {\scriptsize ± 5.80}       & -      & -     \\
     & Baseline + MCSA\textsuperscript{*} & 79.4 {\scriptsize ± 2.92}       & 0.016  & 1.782 \\
     & Baseline + TR\textsuperscript{*} & 81.7 {\scriptsize ± 3.07}       & 0.021  & 1.641 \\
     & TRAS\textsuperscript{**}        & {\bf 84.5} {\scriptsize ± 2.33} & 0.008  & 2.175 \\
    \midrule
    \multirow{4}{*}{\shortstack[l]{%
    Penguins                                                                               \\
    DP: 92.9 {\scriptsize ± 1.85}                                                          \\
        EP: 95.8 {\scriptsize ± 1.72}
      }}
     & Baseline                         & 92.3 {\scriptsize ± 2.89}       & -      & -     \\
     & Baseline + MCSA\textsuperscript{*} & 94.2 {\scriptsize ± 1.33}       & 0.083  & 1.148 \\
     & Baseline + TR\textsuperscript{*} & 96.7 {\scriptsize ± 0.88}       & 0.041  & 1.493 \\
     & TRAS\textsuperscript{**}        & {\bf 98.0} {\scriptsize ± 0.73} & 0.008  & 2.449 \\
    \midrule
    \multirow{4}{*}{\shortstack[l]{%
    Biosses                                                                                \\
    DP: 69.9 {\scriptsize ± 2.73}                                                          \\
        EP: 72.2 {\scriptsize ± 1.78}
      }}
     & Baseline                         & 76.1 {\scriptsize ± 1.97}       & -      & -     \\
     & Baseline + MCSA\textsuperscript{*} & 78.4 {\scriptsize ± 1.66}       & 0.043  & 1.461 \\
     & Baseline + TR\textsuperscript{*} & 83.7 {\scriptsize ± 3.86}       & 0.003  & 3.186 \\
     & TRAS\textsuperscript{**}        & {\bf 88.3} {\scriptsize ± 2.00} & 0.0001 & 8.696 \\
    \midrule
    \multirow{4}{*}{\shortstack[l]{%
    CB                                                                                     \\
    DP: 79.3 {\scriptsize ± 1.57}                                                          \\
        EP: 80.3 {\scriptsize ± 1.54}
      }}
     & Baseline                         & 78.7 {\scriptsize ± 2.13}       & -      & -     \\
     & Baseline + MCSA\textsuperscript{*} & 82.7 {\scriptsize ± 2.31}       & 0.029  & 1.659 \\
     & Baseline + TR\textsuperscript{*} & 85.3 {\scriptsize ± 4.49}       & 0.014  & 2.047 \\
     & TRAS\textsuperscript{**}        & {\bf 87.5} {\scriptsize ± 3.56} & 0.006  & 2.713 \\
    \bottomrule
  \end{tabular}
  \caption{
    APM from \texttt{GPT-3.5-turbo} to \texttt{GPT-4o} (accuracy mean $\pm$ std., five seeds).
    We report DP/EP initial accuracy and final \texttt{GPT-4o} accuracy after optimization (all start from EP).
    Baseline: PromptAgent; +MCSA: aggregation ($K{=}2$); +TR: Textual Regularization from iter.~0; TRAS: both.
    Significance vs.\ baseline via paired $t$-tests ($p$, Cohen's $d$; $^{*}p<0.05$, $^{**}p<0.01$).
    TRAS achieves the best final accuracy ($2.2$\% to $16.0$\% gain).
    }
  \label{tab:prompt_migration_results}
\end{table}

\subsection{Ablation Studies}
\label{sec:ablation}

\begin{figure}[tb]
  \centering
  \begin{subfigure}[t]{0.49\textwidth}
    \includegraphics[width=\linewidth]{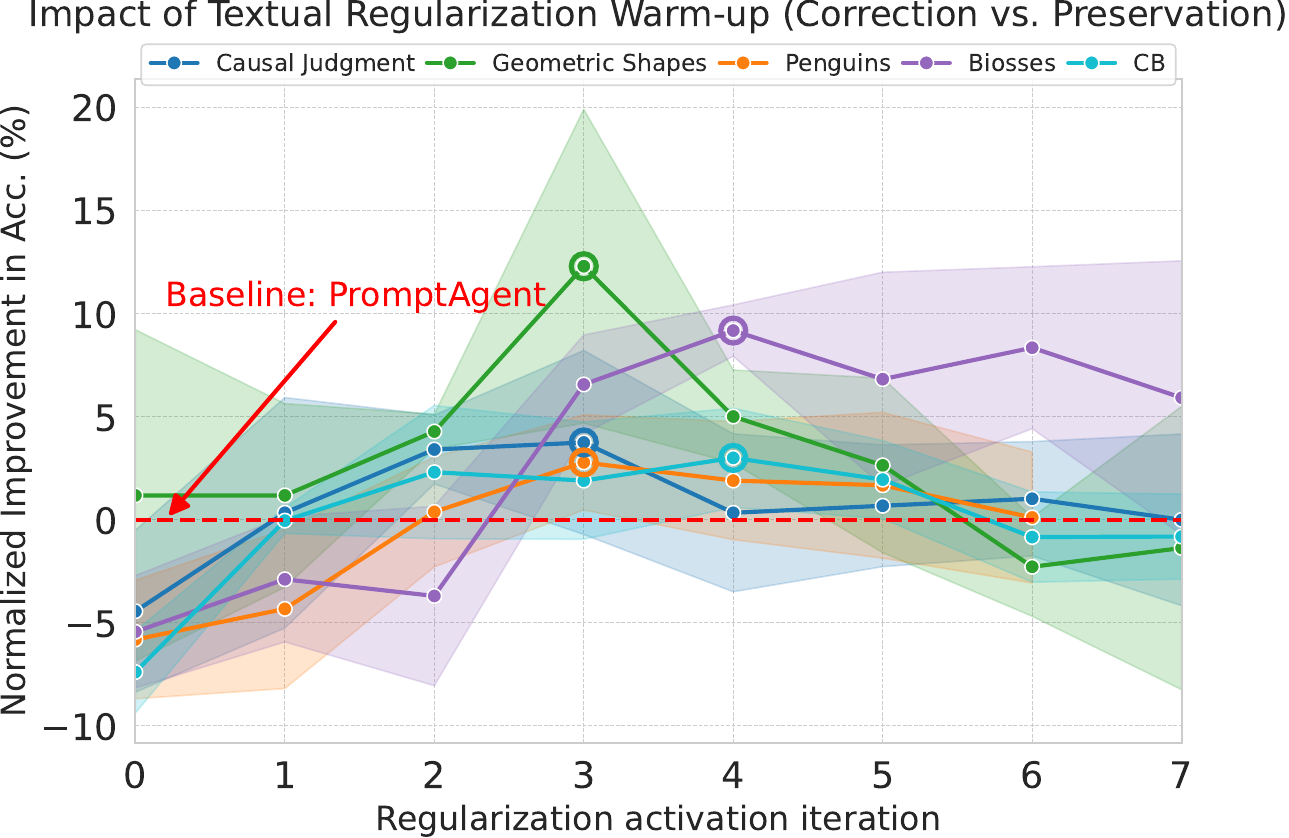}
    \caption{
    Relative accuracy gain vs.\ correction-only baseline as a function of the textual-regularization warm-up $\tau_{\text{warmup}}$.
    Performance peaks at $\tau_{\text{warmup}}\approx 3$--$4$; too early anchors weak prompts, too late increases drift/instruction loss.
    }
    \label{fig:positive_reinforcement_depth}
  \end{subfigure}\hfill
  \begin{subfigure}[t]{0.49\textwidth}
    \includegraphics[width=\linewidth]{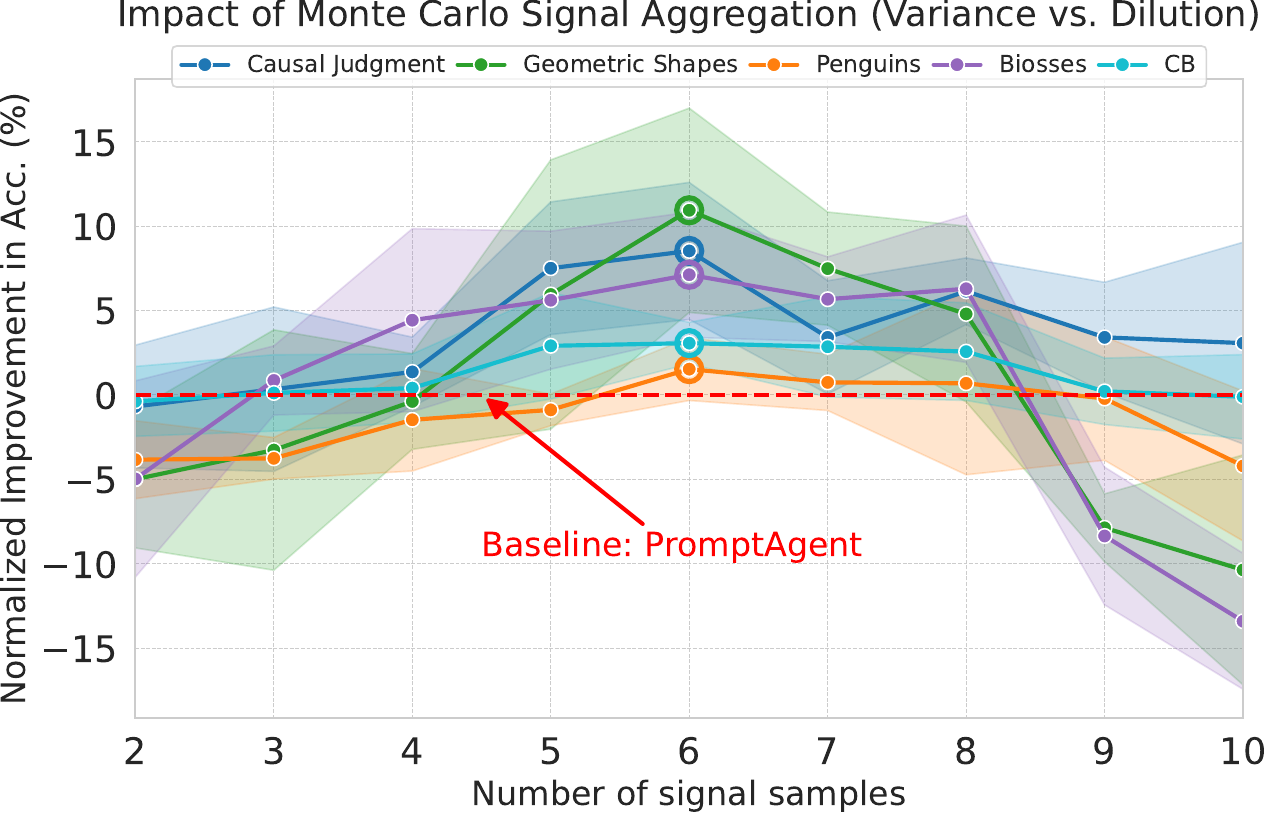}
    \caption{
    Relative accuracy gain vs.\ no-MCSA baseline as a function of aggregation samples $K$.
    Accuracy improves up to $K\approx 6$ (lower variance) then drops for larger $K$ due to dilution/compression.
    }
    \label{fig:feedback_diversification}
  \end{subfigure}
\end{figure}

All ablations use \texttt{GPT-3.5-turbo}, run for 8 iterations from task-specific default prompts (Appendix~\ref{sec:appendix-base-prompt}), and report relative accuracy improvements over the PromptAgent baseline (dotted red line in Figures~\ref{fig:positive_reinforcement_depth} and~\ref{fig:feedback_diversification}).

\textbf{Ablation 1: Textual regularization warm-up ($\tau_{\text{warmup}}$) \quad}
We vary $\tau_{\text{warmup}}$ from 0 to 7 (Figure~\ref{fig:positive_reinforcement_depth}).
Early activation ($\tau_{\text{warmup}}\in\{0,1\}$) underperforms, consistent with premature anchoring to weak prompt structure.
Peak performance occurs at $\tau_{\text{warmup}}=3$ for Causal Judgment, Geometric Shapes, and Penguins, and at $\tau_{\text{warmup}}=4$ for CB and Biosses.
Beyond these values, performance declines as delayed preservation increases semantic drift.

\textbf{Ablation 2: MCSA aggregation samples ($K$) \quad}
We apply MCSA to the baseline using only the corrective signal (Figure~\ref{fig:feedback_diversification}).
Performance improves steadily up to $K\approx 6$, then degrades for larger $K$ (e.g., $8$--$10$) as dilution from over-compression produces generic summaries.
These findings validate the variance--dilution trade-off and motivate $K=6$ in our main APO experiments.

\subsection{Experimentation Costs}
\label{sec:appendix-experimentation-costs}

Table~\ref{tab:call_efficiency} reports the average number of API calls required by each method during prompt optimization (APO) and migration (APM) for both \texttt{GPT-3.5-turbo} and \texttt{GPT-4o}, averaged over five seeds per task.
Methods employing MCSA incur additional calls per iteration due to multiple signal samples, but improved signal quality reduces the number of iterations needed and avoids inefficient ``remove-and-rediscover'' cycles.
In APO, TRAS often reduces total API calls relative to PromptAgent (our default search backbone) by converging more reliably.
In APM, where instruction loss is a primary failure mode, efficiency gains are more pronounced: TRAS reduces total calls by $4.2\%$--$6.2\%$ while also improving accuracy.

\begin{table}[!tb]
  \centering
  \resizebox{\textwidth}{!}{%
  \begin{tabular}{@{} l l r r r r r @{}}
    \toprule
    \textbf{Model}
     & \textbf{Method}
     & \textbf{\shortstack{Causal\\Judgment}}
     & \textbf{\shortstack{Geometric\\Shapes}}
     & \textbf{Penguins}
     & \textbf{Biosses}
     & \textbf{CB} \\
    \midrule
    \multirow{4}{*}{GPT-3.5}
     & Baseline       & 7{,}670  & 11{,}264 & 2{,}664 & 3{,}566 & 5{,}490 \\
     & +MCSA
     & \textcolor{BrickRed}{7{,}919 {\scriptsize(+3.2\%)}}
     & \textcolor{BrickRed}{11{,}731 {\scriptsize(+4.1\%)}}
     & \textcolor{BrickRed}{2{,}735 {\scriptsize(+2.7\%)}}
     & \textcolor{BrickRed}{3{,}710 {\scriptsize(+4.0\%)}}
     & \textcolor{BrickRed}{5{,}606 {\scriptsize(+2.1\%)}} \\
     & +TR
     & \textcolor{OliveGreen}{7{,}040 {\scriptsize(-8.2\%)}}
     & \textcolor{OliveGreen}{10{,}900 {\scriptsize(-3.2\%)}}
     & \textcolor{OliveGreen}{2{,}533 {\scriptsize(-4.9\%)}}
     & \textcolor{OliveGreen}{3{,}294 {\scriptsize(-7.6\%)}}
     & \textcolor{OliveGreen}{5{,}337 {\scriptsize(-2.8\%)}} \\
     & TRAS
     & \textcolor{OliveGreen}{7{,}429 {\scriptsize(-3.1\%)}}
     & \textcolor{OliveGreen}{11{,}204 {\scriptsize(-0.5\%)}}
     & \textcolor{OliveGreen}{2{,}622 {\scriptsize(-1.5\%)}}
     & \textcolor{OliveGreen}{3{,}449 {\scriptsize(-3.3\%)}}
     & \textcolor{OliveGreen}{5{,}422 {\scriptsize(-1.2\%)}} \\
    \midrule
    \multirow{4}{*}{GPT-4o}
     & Baseline       & 8{,}576  & 9{,}642  & 2{,}980 & 3{,}008 & 6{,}229 \\
     & +MCSA
     & \textcolor{BrickRed}{9{,}271 {\scriptsize(+8.1\%)}}
     & \textcolor{BrickRed}{10{,}093 {\scriptsize(+4.7\%)}}
     & \textcolor{BrickRed}{3{,}068 {\scriptsize(+2.9\%)}}
     & \textcolor{BrickRed}{3{,}070 {\scriptsize(+2.0\%)}}
     & \textcolor{BrickRed}{6{,}394 {\scriptsize(+2.6\%)}} \\
     & +TR
     & \textcolor{OliveGreen}{7{,}963 {\scriptsize(-7.1\%)}}
     & \textcolor{OliveGreen}{9{,}396 {\scriptsize(-2.6\%)}}
     & \textcolor{OliveGreen}{2{,}909 {\scriptsize(-2.4\%)}}
     & \textcolor{OliveGreen}{2{,}869 {\scriptsize(-4.6\%)}}
     & \textcolor{OliveGreen}{6{,}079 {\scriptsize(-2.4\%)}} \\
     & TRAS
     & \textcolor{OliveGreen}{8{,}041 {\scriptsize(-6.2\%)}}
     & \textcolor{OliveGreen}{9{,}049 {\scriptsize(-6.1\%)}}
     & \textcolor{OliveGreen}{2{,}825 {\scriptsize(-5.2\%)}}
     & \textcolor{OliveGreen}{2{,}860 {\scriptsize(-4.9\%)}}
     & \textcolor{OliveGreen}{5{,}965 {\scriptsize(-4.2\%)}} \\
    \bottomrule
  \end{tabular}%
  }
  \caption{
Average number of API calls for prompt optimization (APO on GPT-3.5) and migration (APM on GPT-4o), averaged over five runs.
\textcolor{OliveGreen}{Green}/\textcolor{BrickRed}{red} = fewer/more calls vs.\ baseline.
+MCSA: Monte Carlo Signal Aggregation $K{=}6$ (APO) / $K{=}2$ (APM);
+TR: Textual Regularization; TRAS: both.
}
  \label{tab:call_efficiency}
\end{table}

\section{Conclusion and Future Work}
\label{sec:conclusion}

We presented \textbf{TRAS}, a feedback-centric framework that stabilizes black-box APO by augmenting standard textual gradients with \emph{textual regularization} from successes and \emph{Monte Carlo Signal Aggregation} to reduce signal variance.
We also formalized \emph{Automatic Prompt Migration (APM)} as adapting expert prompts across model distributions while avoiding instruction loss.
Across both APO and APM scenarios, TRAS consistently improves accuracy, reduces variance, accelerates convergence, and lowers API usage.
Promising future directions include alternative aggregation mechanisms (e.g., structured voting or embedding-space clustering), adaptive regularization schedules, and extension to multi-turn or continual deployment settings.

\section*{Acknowledgements}
We acknowledge funding from the NSERC Discovery Grant RGPIN-2021-04104 and FRQNT New Scholar. This research was enabled in part by compute resources provided by Digital Research Alliance of Canada (the Alliance) and Calcul Qu\'ebec.

\printbibliography[heading=subbibintoc]

\newpage
\appendix

\section{Dataset}
\label{sec:appendix-dataset-statistics}

We evaluate TRAS on five tasks spanning causal, spatial, tabular, inferential, and semantic reasoning, covering both classification and regression settings.
Three tasks are drawn from BBH~\cite{suzgun2022challenging}, a widely used benchmark in prompt optimization~\cite{schnabel2024sammo,wang2023promptagent,zhou2022large,pryzant2023automatic,yang2023large}:
\textbf{Causal Judgment} (binary causal inference), \textbf{Geometric Shapes} (multi-class reasoning over SVG path strings), and \textbf{Penguins} (table-based classification over structured data).
We additionally include two non-BBH benchmarks:
\textbf{CommitmentBank (CB)}~\cite{de2019commitmentbank}, a natural language inference dataset (entailment/contradiction/neutral),
and \textbf{Biosses}~\cite{souganciouglu2017biosses}, a biomedical semantic similarity benchmark (regression over continuous similarity scores).
Table~\ref{tab:dataset_statistics} summarizes the dataset splits.

\begin{table}[tb]
  \centering
  \setlength{\tabcolsep}{6pt}
  \renewcommand{\arraystretch}{1.0}
  \begin{tabular}{ l c c c }
    \toprule
    \textbf{Dataset} & \textbf{Train} & \textbf{Validation} & \textbf{Test} \\
    \midrule
    Causal Judgment  & 30             & 60                  & 100           \\
    Geometric Shapes & 50             & 100                 & 200           \\
    Penguins         & 30             & 40                  & 79            \\
    Biosses          & 30             & 30                  & 40            \\
    CB               & 30             & 95                  & 56            \\
    \bottomrule
  \end{tabular}
    \caption{Dataset splits used for prompt optimization and evaluation.
  }
  \label{tab:dataset_statistics}
\end{table}

\textbf{Causal Judgment \quad}
This dataset tests causal attribution skills by presenting real-world scenarios and asking whether one event caused another. It evaluates the model's ability to perform commonsense reasoning under ambiguity. Below is an example instance from the dataset:
\begin{examplebox}
  Joe was feeling quite dehydrated, so he stopped by the local smoothie shop to buy the largest sized drink available. Before ordering, the cashier told him that the Mega-Sized Smoothies were now one dollar more than they used to be. Joe replied, ``I don't care if I have to pay one dollar more, I just want the biggest smoothie you have.'' Sure enough, Joe received the Mega-Sized Smoothie and paid one dollar more for it. Did Joe intentionally pay one dollar more?
  \\Label: Yes
\end{examplebox}

\textbf{Geometric Shapes \quad}
A synthetic visual-reasoning-inspired dataset that describes a geometric shape via its SVG representation. The task tests the model's ability to infer spatial and comparative relationships. An example is shown below:
\begin{examplebox}
  This SVG path element <path d=\"M 59.43,52.76 L 75.49,27.45 L 54.92,4.40 M 54.92,4.40 L 23.70,7.77 L 15.15,42.15 L 34.51,57.44 L 59.43,52.76\"/> draws a
  \\Label: hexagon
\end{examplebox}

\textbf{Penguins \quad}
This dataset examinsines the model's ability to reason about tabular data. At each instance, the model is presented with a question about penguins in a table format, and it must select the correct answer from a set of choices.
\begin{examplebox}
  Here is a table where the first line is a header and each subsequent line is a penguin:\\
  \\
  name, age, height (cm), weight (kg)\\
  Louis, 7, 50, 11\\
  Bernard, 5, 80, 13\\
  Vincent, 9, 60, 11\\
  Gwen, 8, 70, 15\\
  For example: the age of Louis is 7, the weight of Gwen is 15 kg, the height of Bernard is 80 cm.\\
  Question: What is the height of Gwen?\\
  Options: A) 50 cm, B) 80 cm, C) 70 cm, D) 60 cm\\
  \\Label: C) 70 cm
\end{examplebox}

\textbf{Biosses \quad}
This biomedical sentence similarity dataset presents pairs of scientific statements and asks for semantic similarity on a scale, assessing the model's ability to reason about specialized, domain-specific language. A sample pair is shown below:
\begin{examplebox}
  S1: It has recently been shown that Craf is essential for Kras G12D-induced NSCLC.\\
  S2: It has recently become evident that Craf is essential for the onset of Kras-driven non-small cell lung cancer.\\
  \\ Label: Similar
\end{examplebox}

\textbf{CB (CommitmentBank) \quad}
CB is a SuperGLUE~\cite{wang2019superglue} NLI dataset designed to evaluate pragmatic inference and speaker commitment in naturally occurring sentences. It differs from standard NLI datasets because each hypothesis is derived directly from the premise's embedded clause, minimizing annotation artifacts. Below is a representative example:

\begin{examplebox}
  Premise: Some of them, like for instance the farm in Connecticut, are quite small. If I like a place I buy it. I guess you could say it's a hobby."\\
  Hypothesis: Buying places is a hobby.\\
  \\Label: Entailment
\end{examplebox}

In this case, the hypothesis is the complement of the clause-embedding verb \texttt{say}, and models must correctly infer that the sentence author is committed to the embedded proposition. This task hinges on understanding modality, clause embedding, and speaker stance, rather than surface-level lexical overlap.

\section{Task Prompts}
\label{sec:appendix-base-prompt}

Each task begins with a minimalist \textit{base prompt} that serves as the initialization point for prompt optimization. These prompts are written as system messages in the \texttt{GPT-3.5-turbo} and \texttt{GPT-4o} chat interfaces, and are intentionally kept simple to avoid embedding task-specific strategies or heuristics. The goal is to provide just enough instruction for the model to attempt the task, allowing the optimization process to refine and expand the prompt effectively.
Below, we list the base prompts used for each task.
\begin{promptbox}{Causal Judgment}
  Answer questions about causal attribution
\end{promptbox}

\begin{promptbox}{Geometric Shapes}
  Name geometric shapes from their SVG paths
\end{promptbox}

\begin{promptbox}{Penguins}
  Answer questions about a table of penguins and their attributes
\end{promptbox}

\begin{promptbox}{Biosses}
  Decide if these two sentences are (A) Not similar (B) Somewhat similar (C) Similar.
\end{promptbox}

\begin{promptbox}{CB}
  What is the relationship between the following premise and the hypothesis?\\
  Options:\\
  - Contradiction\\
  - Neutral\\
  - Entailment
\end{promptbox}

In Section~\ref{sec:prompt_tuning}, we described how each base prompt is optimized. Below, we include the resulting \textbf{expert prompts} obtained from the final iteration of the standard prompt optimization process. These reflect the outcome of the optimization process when targeting performance on the \texttt{GPT-3.5} model.

\begin{promptbox}{Causal Judgment}
  Provide causal attributions in complex scenarios by guiding the model to thoroughly analyze the critical steps, individual intentions, and specific actions that lead to outcomes. Emphasize the importance of identifying and prioritizing the primary cause in each scenario, focusing on direct causes rather than incidental factors. Define clear criteria for evaluating factors and determining the primary cause, considering the combined impact of multiple factors working in conjunction. Instruct the model to weigh the influence of various factors and explicitly guide it on handling conflicting actions and scenarios involving multiple individuals. Ensure that the model carefully considers all significant actions, intentions, and sequences of events leading to the final outcome to accurately attribute causation. Provide explicit instructions for distinguishing between direct causes and incidental factors, prioritizing immediate actions that directly influence outcomes. Define specific criteria for evaluating factors and determining the primary cause, especially in scenarios involving multiple individuals. Emphasize the need to analyze critical steps and actions leading to outcomes in order to accurately attribute causation.
\end{promptbox}
\begin{promptbox}{Geometric Shapes}
  Name the geometric shape accurately based on the provided SVG path. Carefully analyze the properties of the path, including the number of sides, angles, lengths of sides, and overall configuration, to determine the most appropriate geometric shape. Your options should encompass a wide variety of shapes, ranging from simple polygons to circles. Ensure that the model considers all relevant attributes before selecting the most suitable shape from the available options.\\
  Options: (A) circle, (B) equilateral triangle, (C) regular hexagon, (D) rhombus, (E) line segment, (F) octagon, (G) pentagon, (H) rectangle, (I) sector, (J) square, (K) trapezoid, (L) oval
\end{promptbox}

\begin{promptbox}{Penguins}
  Answer questions regarding the following tables of penguins and giraffes, ensuring to accurately reflect any changes made to the penguin table throughout our discussion. Please note these modifications specifically when determining key attributes such as age, weight, or when making comparisons between penguins and giraffes.\\
  \\
  \textbf{Penguin Table:}\\
  name, age, height (cm), weight (kg)\\
  Louis, 7, 50, 11\\
  Vincent, 9, 60, 11\\
  Gwen, 8, 70, 15\\
  \textit{(Any additions or deletions of penguins will be noted in subsequent questions)}\\
  \\
  \textbf{Giraffe Table:}\\
  name, age, height (cm), weight (kg)\\
  Jody, 5, 430, 620\\
  Gladys, 10, 420, 590\\
  Marian, 2, 310, 410\\
  Donna, 9, 440, 650\\
  \\
  For each question, provide clear and logical reasoning behind your answer. Remember to validate the latest state of the penguin table before responding, especially when involving comparisons with giraffes or assessing the attributes of the penguins.\\
  \\
  Additionally, if modifications were made to the penguin table, please annotate them clearly in your response. This ensures that we maintain an accurate understanding of the current data.
\end{promptbox}

\begin{promptbox}{Biosses}
  Decide if these two sentences are (A) Not similar (B) Somewhat similar (C) Similar. Compare the specific regulatory mechanisms and molecular pathways mentioned in each sentence to determine their similarity, explicitly identifying the role of miRNA expression and binding, as well as the relevance of the molecular characteristics of GEFs and nucleotide-binding pockets in the context of the sentences. Analyze both the similarities and differences between the sentences, focusing on the nuances of the regulatory mechanisms and molecular pathways mentioned, and considering the implications for cancer types and cellular processes
\end{promptbox}

\begin{promptbox}{CB}
  What is the relationship between the following premise and the hypothesis?
  Premise: As the storm raged outside, with thunder clapping and lightning illuminating the dark sky, Sarah felt a wave of panic wash over her. She could hear the wind howling, and every crash of thunder made her heart race faster. Despite being tucked away under her thick blankets, she couldn't shake the feeling of terror that gripped her. The flickering candle nearby offered little comfort as she lay wide awake, listening to the chaos around her.\\
  Hypothesis: Sarah felt a strong fear of the storm.\\
  Entailment: The hypothesis is entailed by the premise. Sarah's panic and terror at the storm directly imply that she felt a strong fear of it.
  What is the relationship between the following premise and the hypothesis?\\
  Options:\\
  - Contradiction\\
  - Neutral\\
  - Entailment\\
\end{promptbox}

\end{document}